\documentclass[12pt,a4paper]{article}

\usepackage[british]{babel}

\usepackage[a4paper,top=2cm,bottom=2cm,left=2.5cm,right=2.5cm,marginparwidth=1.75cm]{geometry}

\usepackage[style=apa, backend=biber]{biblatex} 
\DeclareLanguageMapping{british}{british-apa} 
\DeclareFieldFormat[article]{volume}{\apanum{#1}} 

\usepackage{threeparttable}
\usepackage{booktabs}
\usepackage{tabularx, booktabs}
\usepackage{amsmath}
\usepackage{graphicx}
\usepackage[colorlinks=true, allcolors=blue]{hyperref}

\usepackage[title]{appendix}
\usepackage{mathrsfs}
\usepackage{amsfonts}
\usepackage{booktabs} 
\usepackage{caption}  
\usepackage{threeparttable} 
\usepackage{algorithm}
\usepackage{algorithmicx}
\usepackage{algpseudocode}
\usepackage{listings}
\usepackage{enumitem}
\usepackage{chngcntr}
\usepackage{booktabs}
\usepackage{lipsum}
\usepackage{subcaption}
\usepackage{authblk}
\usepackage[T1]{fontenc}    
\usepackage{csquotes}       
\usepackage{diagbox}

\usepackage{setspace}
\usepackage{titlesec}
\titleformat{\section} 
  {\normalfont\Large\bfseries}{\thesection.}{1em}{}
  
\usepackage{lineno} 

\rightlinenumbers 

\linenumbers 

\usepackage{float}   
\usepackage{caption} 


\title{PointABM:Integrating Bidirectional State Space Model with Multi-Head Self-Attention for Point Cloud Analysis}

\author[1]{Jia-wei Chen}
\author[2,*]{Yu-jie Xiong}
\author[3]{Yong-bin Gao}
\affil[1,2,3]{\small The School of Electronic and Electrical Engineering, Shanghai University of Engineering Science, Shanghai 201620, PR China}
\affil[*]{Corresponding author: \texttt{xiong@sues.edu.cn(Y, Xiong)}}

\date{}  

\begin{document}
\maketitle

\begin{abstract}
Mamba, based on state space model (SSM) with its linear complexity and great success in classification provide its superiority in 3D point cloud analysis. Prior to that, Transformer has emerged as one of the most prominent and successful architectures for point cloud analysis. We present PointABM, a hybrid model that integrates the Mamba and Transformer architectures for enhancing local feature to improve performance of 3D point cloud analysis. In order to enhance the extraction of global features, we introduce a bidirectional SSM (bi-SSM) framework, which comprises both a traditional token forward SSM and an innovative backward SSM. To enhance the bi-SSM's capability of capturing more comprehensive features without disrupting the sequence relationships required by the bidirectional Mamba, we introduce Transformer, utilizing its self-attention mechanism to process point clouds. Extensive experimental results demonstrate that integrating Mamba with Transformer significantly enhance the model's capability to analysis 3D point cloud.
\end{abstract}


\section{Introduction}
Point cloud analysis is one of the most widely studied fields of computer vision\parencite{frompoint}. It is widely applications in fields such as autonomous driving and robotics. As a 3D image, point cloud With its own unique data characteristics. It composed of numerous unordered and unpatterned points in three-dimensional space. This necessitates that the entire developmental trajectory of point cloud research be devoted to addressing the challenge posed by the disordered nature of point clouds. To addressing this challenge a variety of method to deep learning have arisen. Vox-based method voxelize the 3D space to enable the application of 3D discrete convolutions \parencite{voxnet}. But this ignore the sparsity of 3D point cloud. \par
Then first work of point-based PointNet \parencite{pointnet} and PointNet++ \parencite{pointnet++} proposed using single symmetric function,max pooling to solving this problem. Subsequently, series point-base models such as PointNeXt \parencite{pointnext}, PointMLP \parencite{pointmlp}, PointCNN \parencite{pointcnn} etc., training form scratch comes out. Transformer-based model\parencite{vistransformer} achieve remarkable progress by its attention mechanism. Attention can effectively capture the relationship between points in point cloud, but also posed quadratic complexity for Transformer \parencite{transformer}. This will cause the increase in model parameters and computational requirements. The permutation invariance of the Transformer endows it with higher compatibility compared to other models. This establishes a foundation for our upcoming proposal to integrate the Transformer and Mamba \parencite{mamba} models.\par
Since Mamba introduction, it has been broadly recognized for its near-linear complexity, drawing substantial attention across various fields. Neverthless, the application of the Mamba model in 3D point clouds still faces many challenges, Mamba model inability fully extract feature information from the point cloud and the scatter and disorder of point clouds cannot meet the requirement of Mamba for sequential order. The results of PointMamBa \parencite{pointmamba} corroborate this.\par
To address this issue, we present our method PointABM. We innovatively combine Mamba model and Transformer model within a novel method. The Transformer model has the characteristics of not changing the arrangement of input elements and having a base, which provides conditions for their combination. PointABM successfully maintains the lightweight characteristics of the Mamba model while effectively leveraging the powerful feature processing capabilities of the Transformer's self-attention mechanism. And we adopted a masked autoencoder pre-training strategy similar to PointMAE, and our method demonstrated exceptional adaptability to this approach. \par

\section{Related work}\label{sec2}

\subsection{Point Cloud Transformers}\label{subsec1}
After the debut of the Point Cloud Transformer \parencite{pct} (PCT), Transformer \parencite{transformer} has continued to be one of the most commonly used models in point cloud analysis \parencite{pointtransformerv1,pointtransformerv2,pointtransformerv3}. This model leverages the powerful attention mechanism of Transformers to better capture the complex spatial relationships in point clouds. The success of PCT demonstrated how to handle the unordered nature of point cloud data through self-attention, while effectively extracting information about the relative positions and attributes between points.\par
Subsequently, PointBERT \parencite{pointbert} and PointMAE \parencite{pointmae} each introduced new pre-training methods for point clouds, incorporating self-supervised learning into Transformer architecture. Both models employ strategy of randomly masking portions of point cloud, significantly enhancing their capability to process and understand the intricate features of point cloud data. These two methods provide stable and reliable pre-training strategies for subsequent models, thereby reducing the reliance on extensive labeled datasets.\par
The exceptional performance of Transformers makes them highly suitable for integration into autoencoders, substantially enhancing downstream point cloud analysis tasks. However, the attention mechanism's \textit{O($n^2d$)} time complexity, with \textit{n} as the input token sequence length and \textit{d} as the Transformer dimension, leads to substantial computational challenges as the input size grows, limiting their efficiency.
 \subsection{State Space Models}\label{subsec2}
State Space Models (SSM), inspired by continuous systems, have emerged as promising frameworks for modeling sequential data. The Structured State Space Sequence Model (S4) \parencite{s4}, a predecessor in this field, is notable for capturing long-range dependencies with linear complexity and strong performance across various domains. To mitigate computational burdens, methods like HTTYH\parencite{HTTYH}, DSS \parencite{DSS}, and S4D \parencite{S4D} employ diagonal matrices within S4. Building on S4, the newly proposed S6 model introduces significant advancements in efficiency and scalability. Mamba \parencite{mamba} further enhances this by introducing selective SSM mechanism, achieving linear-time inference and effective training through hardware-aware algorithm. This innovation has extended to various domains, inspiring works in graph modeling, medical segmentation, and video understanding.\par
Building on S4, the newly proposed S6 model introduces significant advancements, further improving efficiency and scalability. Mamba, which advances this field, introduces a selective SSM mechanism, achieving linear-time inference and effective training through a hardware-aware algorithm. \par
PointMamba \parencite{pointmamba} is the first to introduce the Mamba model into the field of point cloud analysis. However, it does not address the issue of the Mamba model's inability to effectively aggregate local features. Therefore, this paper focuses on combining Transformers with Mamba to leverage the strengths of both.\par

\section{POINTABM}\label{sec3}
\begin{figure}[!ht]
\centering
\includegraphics[width=1\linewidth]{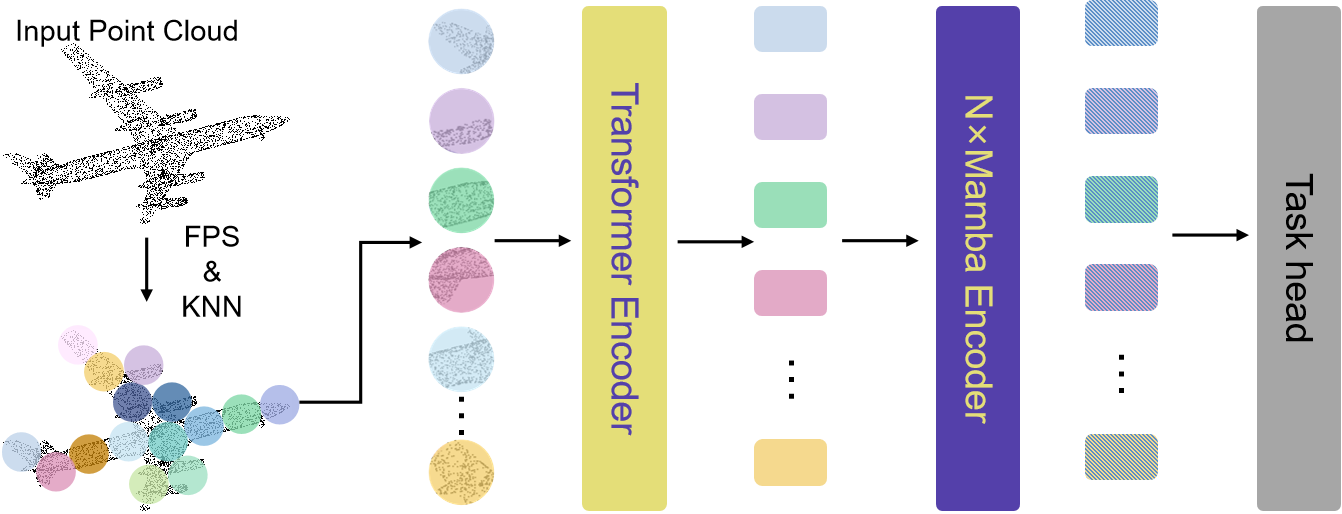}
\caption{\label{fig:cat} The pipeline of PointABM. We initially employ FPS and KNN to extract keypoints and segment them into patches from the input point cloud. Then sent them into Transformer Encoder. Finally, the encoded features are loaded into a Mamba Encoder composed of N bidirectional Mambas.}
\end{figure}

\subsection{Overall}\label{subsec3}
The success of Transformer \parencite{transformer} in various domains \parencite{vistransformer} has proven its power, and its challenger Mamba \parencite{mamba} is constantly refreshing the Sota in various fields \parencite{vismamba,zhou2024mambainmamba,tang2024rotate}.\par
Our method aims to harness the strengths of both Transformer and Mamba in the 3D point clouds domain. To this end, we device a specializing Transformer block to achieve the integration of both models during the manipulating of 3D point clouds. We will introduce multi-head self-attention,  bidirectional Mamba, and the key design of our method.

\subsection{Transformer Block}\label{subsec4}
We employ a standard Transformer composed of multi-head self-attention blocks and feed-forward network (FFN) blocks. The process is shown in Figure 1(a). After resorting, positional encoding is assigned to the features of each center point. 
\begin{equation}
X' = X + \left( \text{Pos} \cdot P + \sum_{i=1}^n \alpha_i \phi_i(\text{Pos}, P) \right)
\end{equation}
The encoded features are segmented and fed into individual self-attention heads. For each head, the input features are multiplied by three learnable weight matrices: $W_Q,W_K,W_V$. \par
\begin{equation}
Q = W_Q X';\quad K = W_K X';\quad V = W_V X'.
\end{equation}
 The $Q, K, V$ matrices undergo self-attention processing.\par
 \begin{equation}
\begin{split}
\text{Attention}(Q,K,V) = \text{softmax} \left( \frac{QK^T}{\sqrt{d_k}} \right) V 
\end{split}
\end{equation}
 Subsequently, the processed features are then combined, rejoined to the original features through a residual connection, and normalized. The introduction of self-attention also enhances our model's adaptability to pre-training methods based on masked autoencoder.\par
  \begin{figure}[!ht]
    \begin{subfigure}{0.4\textwidth}
        \includegraphics[width=\linewidth]{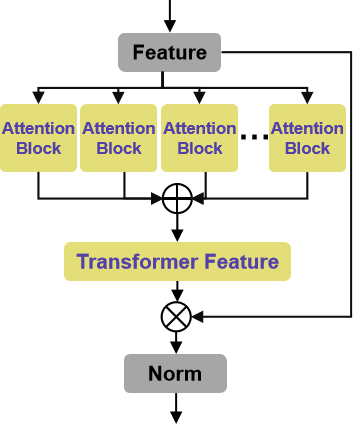}
        \caption{}
    \end{subfigure}
    \hfill
    \begin{subfigure}{0.4\textwidth}
        \includegraphics[width=\linewidth]{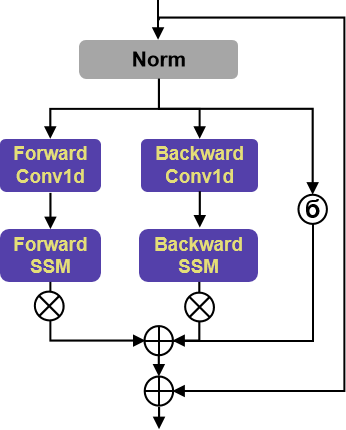}
        \caption{}
    \end{subfigure}
    
    \caption{(a) Transformer Block,\ \ \ \ \ \ \ \ \ \ \ \ \ \ \ \ \ \ \ \ \ \ \ \ \ \ \ \ (b) Bidirectional State Space Block.}
    \label{fig:multi_figs}
\end{figure}
\subsection{Bidirectional State Space Block}\label{subsec5}
The original design of the Mamba block was intended for one-dimensional sequence prediction, which leads to a lack of understanding of the global spatial information required for Point Clouds.To address this issue, we introduced naive Bidirectional State Space Block, which process prediction with forward and backward SSM. \par
We utilize the backward SSM and forward SSM possessed by bi-SSM to process point cloud features and combine the two through a residual connection to strengthen the forward and backward relationships between points. \par
\begin{equation}
T_l = \text{Linear}\left(\sigma(\mathbf{SSM}_{\text{forward}} + \mathbf{SSM}_{\text{backward}}) + T_{l-1}\right)
\end{equation}

\section{Experiments}\label{sec4}
In this section, we will introduce the specific implementation details of the experiment. Then we evaluated the performance of PointABM on ModelNet and three variants of ScanObjectNN. Finally, we show the results of the ablation study for our model.
\subsection{Implementation Details}\label{subsec6}

To address the issue of varying point cloud resolutions, we divide points into different batches. For example, in Modelnet40 \parencite{modelnet40}, with batch size \textit{B} = 1024, it is divided into \textit{n} = 64 point patches, each containing \textit{s} = 32 points, processed using the KNN algorithm. The PointABM encoder features a combination of one Transformer layer and 12 Bi-SSM layers, each with a feature dimension \textit{C} = 384. Each Transformer block consists of 8 heads. We utilize the AdamW optimizer \parencite{adam} and a cosine learning rate decay strategy. In the pretraining phase, we utilize the ShapeNet dataset \parencite{shapenet}, which contains 51,300 3D models. The rest of the settings are essentially the same as those used for training from scratch.All experiments are conducted using an NVIDIA RTX 4090 GPU.

\subsection{Classfication Tasks}\label{subsec7}

\subsubsection{ScanObjectNN}\label{subsubsec1}

\begin{table}[!ht]
\caption{Object classification on ModelNet40 and ScanObjectNN. We evaluate PointABM on three variants, with PB-T50-RS being the most challenging. Overall accuracy (\%) is reported.\label{tab1}}
\begin{threeparttable}
\tiny 
\renewcommand{\arraystretch}{1.2} 
\begin{tabular*}{\columnwidth}{@{\extracolsep\fill}lccccccc@{\extracolsep\fill}}
\toprule
Methods         & Backbone  &Param.(M)& FLOPs(G)&OBJ-BG&OBJ-ONLY& PB-T50-RS \\
\hline
\multicolumn{7}{c}{\textit{Supervised Learning Only}}\\
\hline
    PointNet\textcite{pointnet}    &  -        & 3.5    & 0.5 & 73.3 & 79.2 & 68.0\\
    PointNet++\textcite{pointnet++}  &  -        & 1.5    & 1.7 & 82.3 & 84.3 & 77.9\\
    PointCNN \textcite{pointcnn}   &  -        & 0.6    & 0.9 &86.1 & 85.5 & 78.5\\
    DGCNN   \textcite{dgcnn}    &  -        & 1.8    & 2.4 &82.8 & 86.2 & 78.1\\
    PRA-Net  \textcite{PRA-Net}   &  -        &        & -   &-    & -    & 81.0\\
    MVTN      \textcite{mvtn}  &  -        & 11.2   & 43.7&-    & -    & 82.8\\
    PointNeXt  \textcite{pointnext} &  -        & 1.4    & 1.6 &-    & -    & 87.7\\
    PointMLP   \textcite{pointmlp} &  -        & 12.6   & 31.4&-    & -    & 85.4\\
    DeLA   \textcite{dela} &  -        & 5.3   & 1.5&-    & -    & 90.4\\
\hline
\multicolumn{7}{c}{\textit{Training from pre-training}}\\
\hline
Point-BERT  \textcite{pointbert}    &Transformer\textcite{transformer}& 22.1   & 4.8 & 87.43 & 88.12 & 83.07\\
MaskPoint   \textcite{MaskPoint}   &Transformer& 22.1   & 4.8 & 89.30 & 88.10 & 84.30\\
Point-MAE    \textcite{pointmae}   &Transformer& 22.1   & 4.8 &90.02 & 88.29 & 85.18\\
Point-M2AE    \textcite{pointm2ae}   &Transformer& 15.3   & 3.6 &91.22 & 88.29 & 85.18\\
PointMamba     \textcite{pointmamba} &  Mamba\textcite{mamba}    & 12.3   & 3.1 &88.29 & 87.78  & 82.48\\
PointMamba-pre \textcite{pointmamba} &  Mamba    & 12.3   & 3.1 &90.71 & 88.47 & 84.87\\
PCM            \textcite{PCM} &  Mamba    & 34.2   & 45.0 &-     & -     & 88.1\\
PointABM(ours) &Mamba,Transformer & 15.1   & 9.6 &91.57 & 90.36 & 86.19\\
PointABM-pre(ours) &Mamba,Transformer & 15.1   & 9.6& \textbf{93.29}& \textbf{92.43}& \textbf{88.29}\\
\bottomrule
\end{tabular*}
\end{threeparttable}
\end{table}

ScanObjectNN\parencite{scanobjectnn} dataset comprises 15,000 objects segmented into 15 categories, captured from real-world indoor environments characterized by their cluttered backgrounds. This dataset presents three distinct variants for testing and analysis: OBJ\_BG, OBJ\_ONLY, and PB\_T50\_RS, each designed to evaluate different aspects of object recognition under varying complexly conditions. The configuration for our experiments taking a subset of 2,048 points per object and using rotation as data augmentation. \textbf{PointABM} surpasses most effective Transformer-based method  PointMAE, 3.58\%, 4.14\%, 3.42\% on OBJ\_BG, OBJ\_ONLY, and PB\_T50\_RS. Besides, we also exceeding Mamba-based mothod PointMamba 2.58\%, 3.96\%, 3.33\%. 

\subsubsection{ModelNet40}\label{subsubsec2}

\begin{minipage}[t]{.40 \textwidth}
ModelNet40\parencite{modelnet40} is a widely recognized synthetic dataset for 3D object classification, comprising 12,311 clean CAD models across 40 categories. The dataset is conventionally split into 9,843 instances for training and 2,468 for testing, adhering to established protocols. Each category is represented by 100 unique models, establishing ModelNet40 as a fundamental benchmark in the field. During training, random scaling and translation are employed to enhance generalization. Despite its status as a clean dataset, PointABM's inability to fully demonstrate interference resistance still resulted in an impressive accuracy rate of 93.1 \%.

\end{minipage}%
\hfill
\hspace{30pt} 
\begin{minipage}[t]{.60\textwidth} 
    \begin{threeparttable}
        \captionof{table}{Object classification on ModelNet40.\label{tab2}}
        \scriptsize 
        \renewcommand{\arraystretch}{1.2} 
        \begin{tabular}{lcc}
            \toprule
            Methods & Backbone & ModelNet40 (\%) \\
            \midrule
            PointNet & - & 89.2 \\
            PointNet++& - & 90.7 \\
            PointCNN & - & 92.2 \\
            DGCNN & - & 92.9 \\
            PRA-Net & - & 93.1 \\
            MVTN& - & 93.8 \\
            PointNeXt & - & 94.0 \\
            PointMLP & - &  \textbf{94} \\
            DeLA & - & 94.0 \\
            Point-BERT & Transformer & 93.4 \\
            MaskPoint & Transformer & 93.8 \\
            Point-MAE & Transformer & 94.4 \\
            Point-M2AE & Transformer & 94.0 \\
            PointMamba & Mamba & 92.4 \\
            PointMamba-pre & Mamba & 93.6 \\
            PCM & Mamba & 93.4 \\
            PointABM & Mamba, Transformer & 92.6 \\
            PointABM-pre & Mamba, Transformer & 93.1 \\
            \bottomrule
        \end{tabular}
    \end{threeparttable}
\end{minipage}

\subsection{Ablation study}\label{subsec8}
For improving effect of each component, we conducted a study on the utility of each component within the architecture on ScanObjectNN\parencite{scanobjectnn}. And to ensure the purity of the ablation study results, all our ablation experiments were conducted using training from scratch.

\subsubsection{Transformer embedding}\label{subsubsec3}

\begin{table}[ht]
    \centering
    \begin{threeparttable}
        \captionof{table}{the effect of Transformer embedding.\label{tab3}}
        \scriptsize 
        \renewcommand{\arraystretch}{1.2} 
        \begin{tabular}{lccccc}
            \toprule
            fusion method&OBJ\_BG & OBJ\_ONLY & PB\_T50\_RS (\%) &feature dimension&Param.(M)\\
            \midrule
            None & 88.3 & 87.8 &82.5&384&12.3\\
            Concatenation & 90.72 & 88.29&84.62&768&47.7\\
            Residual Connection & 90.43 & 89.15 &84.55&384&14.8\\

            \bottomrule
        \end{tabular}
    \end{threeparttable}
\end{table}
As the first to integrate Transformer and Mamba in the point cloud field, we attempted two feature fusion methods: concatenation and residual connection. Table. \ref{tab3} shows each feature fusion method brought a noticeable improvement. This indicates that Transformer embedding can effectively offer more refined feature information to the Mamba model. Concatenation feature dimension even take better accuracy. But with the doubling of feature dimensions, the size of the model increases dramatically. Moreover, in the subsequent ablation studies of BI-SSM, the feature fusion method using residual connections demonstrated superior compatibility.

\subsubsection{Bidirectional Mamba embedding}\label{subsubsec4}
In this section, we will focus on examining the compatibility of Bidirectional Mamba embedding with two types of Transformer embedding blocks. The introduction of the Transformer block in the previous section, with its attention mechanism, enhances the temporal relationships in point cloud data. Therefore, this chapter does not conduct separate experiments on Bidirectional Mamba. Instead, it tests the outcomes when combined with the two different feature fusion methods discussed in Section 4.3.1.\par
\begin{table}[ht]
    \centering
    \begin{threeparttable}
        \captionof{table}{the effect of Bidirectional Mamba embedding.\label{tab4}}
        \scriptsize 
        \renewcommand{\arraystretch}{1.2} 
        \begin{tabular}{lccccc}
            \toprule
            method&OBJ\_BG & OBJ\_ONLY & PB\_T50\_RS (\%) &feature dimension&Param.(M)\\
            \midrule
            Concatenation & 91.57 & 88.81&84.17&768&48.1\\
            Residual Connection &91.57 & 90.36 &86.19&384&15.1\\

            \bottomrule
        \end{tabular}
    \end{threeparttable}
\end{table}
By using Bidirectional Mamba, we achieved significant improvements in the embedding process. Compared to the simpler Mamba model, PointABM enhanced performance by 1.14\%, 1.21\%, and 1.64\% across three distinct datasets, respectively.

\section*{Acknowledgments}
thanks


\printbibliography


\renewcommand\theequation{\Alph{section}\arabic{equation}} 
\counterwithin*{equation}{section} 
\renewcommand\thefigure{\Alph{section}\arabic{figure}} 
\counterwithin*{figure}{section} 
\renewcommand\thetable{\Alph{section}\arabic{table}} 
\counterwithin*{table}{section} 

\end{document}